\documentclass{article}
\usepackage[main, final]{neurips_2026}
\usepackage[utf8]{inputenc} 
\usepackage[T1]{fontenc}    
\usepackage{hyperref}       
\hypersetup{hidelinks}
\usepackage{url}            
\usepackage{booktabs}       
\usepackage{amsfonts}       
\usepackage{nicefrac}       
\usepackage{microtype}      
\usepackage{needspace}

\usepackage{graphicx}
\usepackage{subcaption}
\usepackage{multirow}
\usepackage{amsmath}
\usepackage{amssymb}
\usepackage{mathtools}
\usepackage{amsthm}
\usepackage[table]{xcolor}
\DeclareRobustCommand{\rev}{\color{black}}
\usepackage{wrapfig}
\usepackage{fontawesome5}
\newcommand{\hflogo}{\raisebox{-0.25\height}{\includegraphics[height=1.45em]{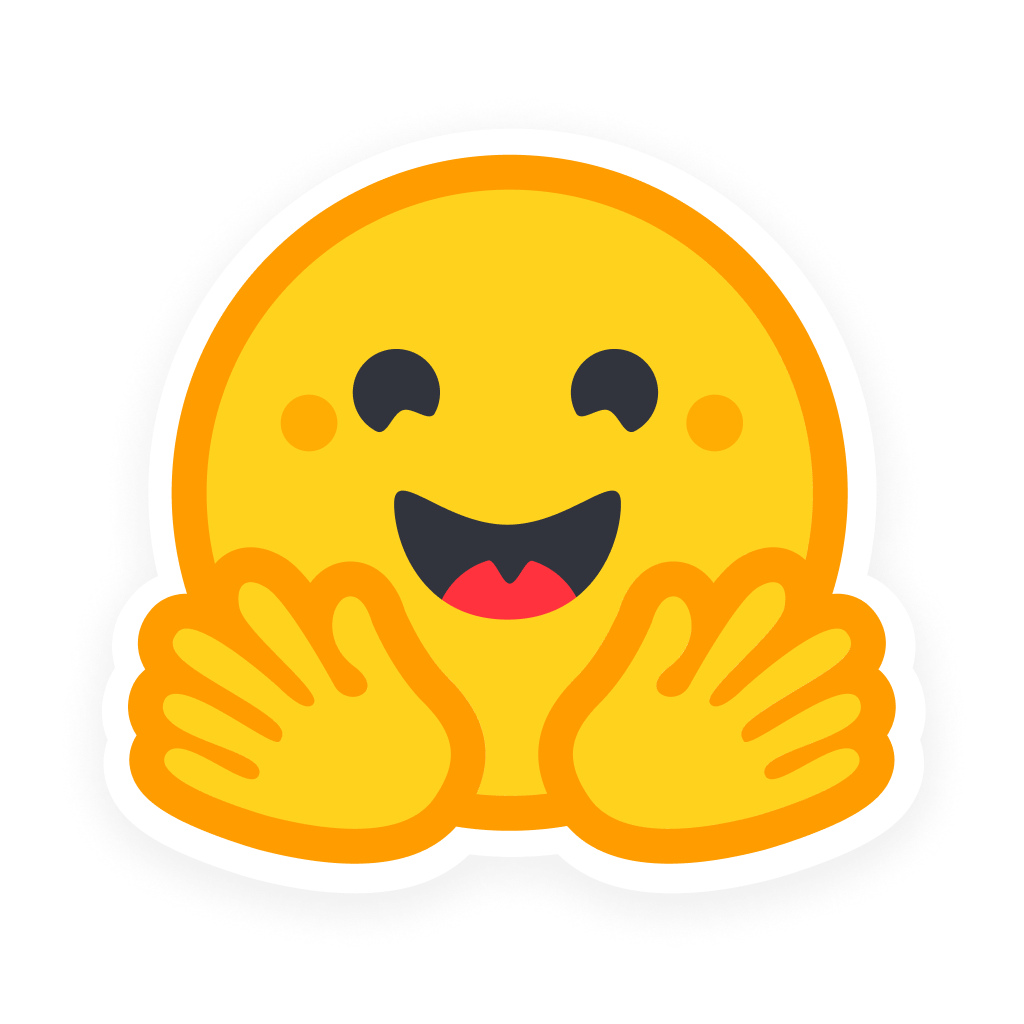}}}
\usepackage{listings}
\lstdefinestyle{myprompt}{
    backgroundcolor=\color{gray!10},    
    basicstyle=\ttfamily\small,         
    frame=single,                       
    framesep=5pt,                       
    breaklines=true,                    
    postbreak=\raisebox{0ex}[0ex][0ex]{\ensuremath{\hookrightarrow\space}},
    showstringspaces=false,              
    breakindent=0pt,                     
    postbreak={},                        
    morekeywords={system_prompt, user_prompt}, 
    keywordstyle=\bfseries,
}

\title{Verifiable Process Rewards for Agentic Reasoning}

\author{%
  Huining Yuan$^{*}$, Zelai Xu$^{*}$, Huaijie Wang, Xiangmin Yi, Jiaxuan Gao, \\
  \textbf{Xiao-Ping Zhang, Yu Wang$^{\dagger}$, Chao Yu$^{\dagger}$, Yi Wu$^{\dagger}$}\\
  Tsinghua University\\
  \vspace{0.5em}
  \href{https://thu-nics.github.io/VPR/}{\textcolor{black}{\faGlobe\ Project Page}} \quad
  \href{https://github.com/thu-nics/VPR}{\textcolor{black}{\faGithub\ Code}} \quad
  \href{https://huggingface.co/collections/nics-efc/vpr}{\textcolor{black}{\hflogo\ Models}}
}

\begin{document}

\begingroup
\let\thefootnote\relax
\footnotetext{$^*$Equal contribution: \texttt{\{yuanhuining0, zelai.eecs\}@gmail.com}}
\footnotetext{$^\dagger$Corresponding authors: \texttt{yu-wang@tsinghua.edu.cn, yuchao@sz.tsinghua.edu.cn, jxwuyi@gmail.com}}
\endgroup

\maketitle

\begin{abstract}
{\color{black}Reinforcement learning from verifiable rewards (RLVR) has improved the reasoning capabilities of large language models (LLMs), yet existing approaches predominantly rely on sparse outcome-level feedback, posing a substantial credit-assignment challenge in long-horizon agentic tasks.
In this work, we study structured agentic reasoning problems that admit symbolic or algorithmic task-solving oracles. We propose Verifiable Process Rewards (VPR), a {\rev training} framework that repurposes such task oracles as action-level process verifiers, providing dense rewards for intermediate actions and enabling fine-grained credit assignment.
We instantiate VPR in three distinct settings: a search-based oracle for Sokoban, a constraint-based oracle for Sudoku, and a posterior-based oracle for Minesweeper. To optimize policies with these rewards, we introduce oracle-guided state-group rollout and locally normalized optimization.}
{\color{black}Empirically, VPR-trained Qwen3-4B outperforms outcome-reward GRPO and VinePPO across all three training environments.
When jointly trained on math data and all three game environments, VPR achieves the {\rev highest average general-reasoning performance and strongest agentic results among the compared training configurations}.
To further assess its applicability to more general agentic tool-use settings, we apply VPR-style action-level supervision to $\tau^2$-bench using an expert LLM as a reference policy and observe improved performance over outcome-reward GRPO on held-out tasks.
Together, these results position VPR as a promising approach for leveraging reliable task oracles to improve agentic reasoning, while highlighting its dependence on the availability and reliability of task oracles and the challenge of extending intermediate verification to less structured, open-ended environments.
}

\end{abstract}

\section{Introduction}
\label{sec:intro}
{\color{black}
Reinforcement learning from verifiable rewards (RLVR) has become an effective approach for improving the reasoning capabilities of large language models (LLMs) by optimizing against objective, task-grounded signals such as exact-answer checks in mathematics and unit-test outcomes in code~\cite{guo2025deepseek,jaech2024openai,shao2024deepseekmath}. Yet most existing methods provide feedback only after a final answer or an entire trajectory has been completed. In long-horizon agentic tasks involving tool use, environment interaction, and multi-turn planning~\cite{yao2022webshop,jimenez2024swebench}, such sparse outcome-level supervision makes it difficult to assign credit to intermediate actions.

Process supervision addresses this ambiguity by assigning feedback to intermediate decisions. Existing Process Reward Models (PRMs)~\cite{lightman2024lets,uesato2022solving} often rely on learned reward models, LLM judges, or Monte Carlo rollouts. Learned or generative feedback can be noisy, biased, or susceptible to reward hacking~\cite{llmasajudge,huang2024large}, whereas rollout-based value estimates require multiple continuations from intermediate states and can be computationally expensive and high variance~\cite{kazemnejad2025vineppo,wang-etal-2024-math}. Dense feedback is therefore most useful when it is also reliable and grounded in the structure of the task.}

{\color{black}In this work, we study structured agentic reasoning problems that admit symbolic or algorithmic task-solving oracles{\rev{} queryable at intermediate states}. Such oracles exploit explicit task structure: search procedures produce solution paths in dynamic environments, constraint solvers construct consistent assignments, and inference procedures derive decisions under uncertainty. Although these oracles encode effective task-solving procedures, their outputs are not directly defined as reward functions over arbitrary actions generated by a learned policy.}

\begin{figure*}[t]
\centering
\includegraphics[width=\linewidth]{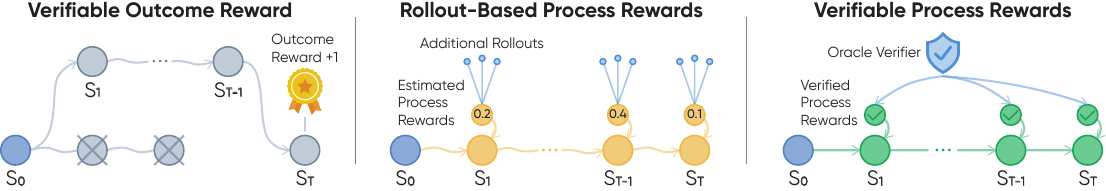}
\caption{{\color{black}Three reward designs for long-horizon reasoning. \textbf{Left:} outcome reward is observed only at the end of a trajectory. \textbf{Middle:} rollout-based process rewards, instantiated by VinePPO in our experiments, estimate intermediate values using additional policy rollouts. \textbf{Right:} VPR repurposes a task-solving oracle as a process verifier that directly scores intermediate actions.}}
\label{fig:overview}
\end{figure*}

{\color{black}We propose \emph{Verifiable Process Rewards} (VPR), a {\rev training} framework that repurposes task-solving oracles as action-level process verifiers for reinforcement learning. Figure~\ref{fig:overview} contrasts VPR with outcome rewards and rollout-based process rewards. At each environment step, VPR queries the task-specific oracle and treats the oracle-selected action, together with any oracle-equivalent alternatives, as the oracle action set. An action receives $+1$ if it belongs to this set, $0$ if it is legal but outside the set, and $-1$ if it is illegal or unparsable. We instantiate this oracle-to-reward construction using breadth-first search (BFS) in \emph{Sokoban}, constraint-based solving in \emph{Sudoku}, and posterior inference in \emph{Minesweeper} (Figure~\ref{fig:vpr}). {\rev The task oracles provide guidance during training; the learned policies are evaluated without oracle assistance.}

To optimize policies with these rewards, we introduce two coupled algorithmic techniques: oracle-guided state-group rollout and locally normalized optimization. At each visited state, the policy samples multiple candidate responses, and the constructed verifier scores and ranks their parsed actions. A single candidate is sampled uniformly from the highest-reward set and committed to the environment. Repeating this best-of-$K$ commit at every step steers the committed trajectory toward promising successor states{\rev{}, directing exploration through an oracle-guided shift in the visited-state distribution}. Meanwhile, all candidate responses sampled from the same environment state form a localized state group, in contrast to the trajectory group in standard GRPO~\cite{shao2024deepseekmath}. Their rewards are then centered within the state group to produce relative advantages, allowing {\rev every informative, non-padding} candidate{\rev{} response} to contribute to the policy update. {\rev Together, the oracle action set, active same-state sampling, local updates, and guided commit form VPR's integrated training mechanism.}
}

{\color{black}We evaluate VPR against standard GRPO and VinePPO~\cite{kazemnejad2025vineppo} in the three training environments. 
In these in-domain evaluations, VPR-trained Qwen3-4B outperforms GRPO and VinePPO on every reported metric across all three environments. 
To further examine whether VPR training fosters generalizable agentic reasoning that transfers beyond these game environments, we train Qwen3-4B-Base jointly on math data and interactions from all three games. 
Out-of-domain (OOD) results show that VPR achieves the highest average performance across seven general-reasoning benchmarks and the strongest results on the agentic ALFWorld and WebShop benchmarks{\rev{} among the compared training configurations}. }

{\color{black}Taken together, our results highlight two insights. First, when a reliable task-solving oracle is available{\rev{} and its guidance can be expressed as an oracle action set}, VPR can repurpose it as an action-level process verifier, providing a dense alternative to sparse outcome-reward training. Second, incorporating VPR supervision from {\rev multiple environments with task oracles can improve performance beyond those training environments in our mixed-training experiments}. VPR nevertheless remains limited by the availability and reliability of task oracles, leaving less structured, open-ended environments as an important challenge. On $\tau^2$-bench, we further examine whether VPR-style action supervision extends to general tool use and whether an expert reference policy can serve as an approximate oracle at student-visited states.} 

Our contributions are summarized as follows:
\begin{itemize}
    \item {\color{black}We introduce \emph{Verifiable Process Rewards} (VPR), {\rev a unified framework coupling oracle-derived action-level rewards with guided rollout and local policy updates}.}

    \item {\color{black}We instantiate VPR with BFS in Sokoban, constraint solving in Sudoku, and posterior inference in Minesweeper.}

    \item {\color{black}We propose oracle-guided state-group rollout and locally normalized optimization for guided exploration and same-state credit assignment.}

    \item {\color{black}We demonstrate consistent in-domain gains and stronger average OOD transfer, conduct ablations, and evaluate expert-as-oracle VPR for general tool use on $\tau^2$-bench.}
\end{itemize}

\section{Method}
\label{sec:method}
{\rev We present VPR as a unified training procedure that combines an oracle action set, active same-state candidate sampling, locally normalized updates, and an oracle-guided commit.}

{\color{black}
\subsection{From Task-Solving Oracles to Process Verifiers}
\label{sec:method:preliminaries}

{\rev We consider episodic agentic reasoning environments with deterministic transitions and potentially partial observations. The underlying process is $(\mathcal{S},\mathcal{A},\mathcal{P},R,H)$}, where $\mathcal P:\mathcal S\times\mathcal A\rightarrow\mathcal S$ is the transition function. {\rev At full environment state $s_t$, the policy receives only the observation $o_t=h(s_t)$.} The LLM samples a complete response $y_t\sim\pi_\theta(\cdot\mid {\rev o_t})$, and the environment action parser maps it to a candidate action $a_t=\psi(y_t)$. For a valid parsed action $a_t\in\mathcal A$, the environment transitions to $s_{t+1}=\mathcal P(s_t,a_t)$; illegal or unparsable outputs are handled by the verifier below. The resulting environment trajectory is $\tau=(s_1,a_1,r_1,\ldots,s_T,a_T,r_T)$, where $T\le H$. Outcome-level RL supplies only the sparse reward
\begin{equation}
    r_t^{\mathrm{OR}}=0 \quad (t<T),
    \qquad
    r_T^{\mathrm{OR}}=\mathbb{I}(\mathrm{success}),
\end{equation}
which is objective but does not identify the intermediate decisions responsible for the result.

We assume that the structured task admits a symbolic or algorithmic task-solving oracle $\mathcal O${\rev{} that can be queried at policy-visited states, together with task knowledge sufficient to identify oracle-preferred and legal actions}. Depending on the task, $\mathcal O$ may return a solution path, a constraint-consistent assignment, or a posterior-based decision rule; its native output need not score arbitrary actions generated by the policy. At state $s$, running the task oracle induces an \emph{oracle action set} $\mathcal A_{\mathcal O}(s)$, containing all actions that are equally preferred under the task-specific oracle criterion. VPR repurposes the oracle as the following action-level process verifier:
\begin{equation}
    r_t^{\mathrm{VPR}}
    =\mathcal V_{\mathcal O}(s_t,a_t)
    =\begin{cases}
        +1, & a_t\in\mathcal A_{\mathcal O}(s_t),\\
        \phantom{+}0, & a_t\text{ is legal and }a_t\notin\mathcal A_{\mathcal O}(s_t),\\
        -1, & a_t\text{ is illegal or invalid}.
    \end{cases}
\end{equation}
Thus, an oracle action receives $+1$, any other legal action receives $0$, and an illegal or unparsable action receives $-1$. {\rev The reward mapping and training procedure are shared across tasks; the oracle action set and legality criterion remain task specific.} The mapping is deterministic and evaluates only the parsed environment action; no learned reward model observes the policy's hidden reasoning trace. This action-level supervision guides the policy toward oracle decisions without directly prescribing the agent's reasoning process.

\begin{figure*}[t]
\centering
\includegraphics[width=\linewidth]{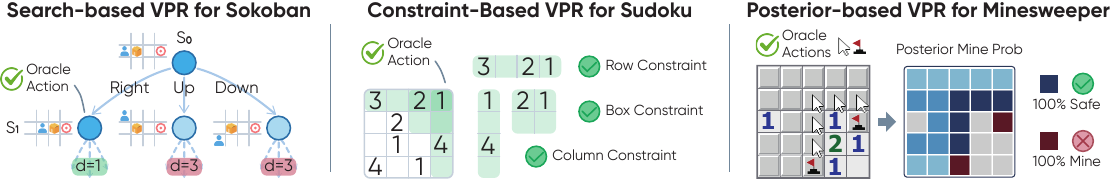}
\caption{{\color{black}Three VPR instantiations of the shared oracle-action-set principle. \textbf{Search-based} (\textit{Sokoban}): BFS solution paths define the oracle action set. \textbf{Constraint-based} (\textit{Sudoku}): the solution and current constraints identify preferred solution-consistent fills. \textbf{Posterior-based} (\textit{Minesweeper}): posterior inference identifies safe reveals, certain flags, and minimum-risk guesses.}}
\label{fig:vpr}
\end{figure*}

\subsection{Three Instantiations of VPR}
\label{sec:method:vpr}

\textbf{Search-Based VPR for Sokoban.}
The task-solving oracle uses bounded breadth-first search (BFS) to find shortest solution paths. For a state solved within the search budget, let $D(s)$ denote its shortest distance to completion. The oracle action set contains every first move on a shortest path,
\begin{equation}
    \mathcal{A}_{\mathcal O}(s)
    =\left\{a\in\mathcal{A}(s):D(\mathcal P(s,a))=D(s)-1\right\},
\end{equation}
where $\mathcal A(s)\subseteq\mathcal A$ is the set of legal actions at $s$. Every action in $\mathcal A_{\mathcal O}(s)$ receives $+1$, any other legal move receives $0$, and a malformed or illegal move receives $-1$. Including every equally short first move avoids imposing an arbitrary tie break from a single returned solution path.

\textbf{Constraint-Based VPR for Sudoku.}
The oracle combines the unique solution $G^\star$ with the current constraints. It selects correct fills of forced cells when available and otherwise correct fills of minimum-remaining-value (MRV) cells; all ties form $\mathcal A_{\mathcal O}(s)$ and receive $+1$. Other correct legal fills receive $0$, while incorrect or malformed actions receive $-1$.

\textbf{Posterior-Based VPR for Minesweeper.}
{\rev Given observation $o=h(s)$, containing the revealed board and global mine count, the oracle performs posterior inference without accessing the actual hidden layout. Let $\Omega(o)$ denote the consistent hidden mine configurations:}
\begin{equation}
    P(\mathrm{mine}_{i,j}\mid {\rev o})
    =\frac{\sum_{\omega\in{\rev\Omega(o)}}\mathbb{I}((i,j)\text{ is a mine in }\omega)}{|{\rev\Omega(o)}|}.
\end{equation}
The oracle prioritizes safe reveals with posterior zero and certain flags with posterior one; when neither exists, it selects minimum-posterior guesses. All actions tied at the highest available priority form $\mathcal A_{\mathcal O}(s)$ and receive $+1$. Other legal reveals or flags receive $0$, and malformed or illegal actions receive $-1$. This construction preserves equivalent actions under posterior uncertainty rather than forcing an arbitrary single choice.
}

{\color{black}
\subsection{State-Group Rollout and Optimization}
\label{sec:method:optimization}

{\rev A \emph{state group} contains candidate responses sampled at one intermediate state. Selected actions form the \emph{committed trajectory}. In contrast, a GRPO \emph{trajectory group} contains complete trajectories from one initial instance and normalizes their outcome rewards.}

{\rev At state $s_g$, the behavior policy independently samples $K$ complete responses conditioned on the shared observation $o_g=h(s_g)$, followed by action parsing and verification:}
\begin{equation}
    y_{g,k}\sim\pi_{\theta_{\mathrm{old}}}(\cdot\mid {\rev o_g}),
    \qquad
    a_{g,k}=\psi(y_{g,k}),
    \qquad
    r_{g,k}=\mathcal{V}(s_g,a_{g,k}),
    \qquad k\in\{1,\ldots,K\}.
\end{equation}
The verifier scores the parsed action rather than the hidden reasoning text. Candidates share the same serialized {\rev observation} while retaining their individual response, action, reward{\rev{}, and metadata}. We use $K=4$ in all state-group experiments.

\begin{wrapfigure}[11]{r}{0.45\textwidth}
\vspace{-11pt}
\centering
\includegraphics[width=\linewidth]{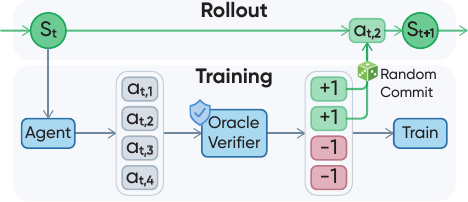}
\caption{{\color{black}State-group rollout and training. One maximum-reward candidate advances the rollout, while every {\rev informative, non-padding} candidate sampled at that state contributes to the policy update.}}
\label{fig:state_group}
\vspace{-8pt}
\end{wrapfigure}
Candidate sampling, policy optimization, and environment transition are distinct operations. After scoring the candidates, we sample one index uniformly at random from the maximizers,
\begin{equation}
    \begin{aligned}
        k_g^\star &\sim \operatorname{Unif}\!\left(\arg\max_{k}r_{g,k}\right), \\
        s'_g &= \mathcal{P}(s_g,a_{g,k_g^\star}).
    \end{aligned}
\end{equation}
Only $a_{g,k_g^\star}$ is committed to the environment, so exactly one successor state $s'_g$ begins the next group. The remaining candidates are not discarded: every eligible candidate contributes to the policy update.

Repeating the maximum-reward commit at every step also serves an exploration role: by making oracle-preferred successor states more likely to be visited, it increases the chance that later state groups are collected along promising trajectories. This mechanism induces an oracle-guided shift in the visited-state distribution relative to ordinary single-sample policy rollouts. {\rev Candidate responses themselves remain sampled from $\pi_{\theta_{\mathrm{old}}}(\cdot\mid o_g)$ at their source state.} Increasing $K$ raises the probability of finding an oracle-preferred action but also strengthens the selection pressure and corresponding state-distribution shift. We therefore use $K=4$ as a practical trade-off{\rev{} between guided exploration and this distribution shift}.

For each group, we first center rewards within the shared source state,
\begin{equation}
    \widetilde A_{g,k}=r_{g,k}-\bar r_g,
    \qquad
    \bar r_g=\frac{1}{K}\sum_{j=1}^{K}r_{g,j}.
\end{equation}
Groups with identical rewards contain no within-state preference signal and are masked out. For the remaining non-padding candidates, the standalone game estimator whitens the centered values across the update batch $\mathcal B_+$,
\begin{equation}
    \label{eq:state-group-batch-whiten}
    A_{g,k}=\frac{\widetilde A_{g,k}-\mu_{\mathcal B_+}}
    {\sigma_{\mathcal B_+}+\delta},
    \qquad (g,k)\in\mathcal B_+.
\end{equation}
Divisibility padding never enters $\bar r_g$, $\mu_{\mathcal B_+}$, $\sigma_{\mathcal B_+}$, metrics, or gradients. To express the implemented token-level update, write response $y_{g,k}=(y_{g,k,1},\ldots,y_{g,k,L_{g,k}})$ and let $\mathcal T_+$ contain all response-token indices $(g,k,i)$ that remain active after applying the candidate, equal-reward-group, and padding masks. The token-level importance ratio is
\begin{equation}
    \rho_{g,k,i}(\theta)
    =\frac{\pi_\theta(y_{g,k,i}\mid {\rev o_g},y_{g,k,<i})}
    {\pi_{\theta_{\mathrm{old}}}(y_{g,k,i}\mid {\rev o_g},y_{g,k,<i})}.
\end{equation}
The candidate-level advantage $A_{g,k}$ is broadcast to every active token in its response, and the clipped surrogate uses global token-mean aggregation,
\begin{equation}
\begin{aligned}
    J_{\mathrm{VPR}}(\theta)
    =\frac{1}{|\mathcal T_+|}
    \sum_{(g,k,i)\in\mathcal T_+}\!\min\!\left(
      \rho_{g,k,i}(\theta)A_{g,k},
      \operatorname{clip}\!\left(\rho_{g,k,i}(\theta),1-\epsilon_{\mathrm{clip}},1+\epsilon_{\mathrm{clip}}\right)A_{g,k}
    \right).
\end{aligned}
\end{equation}
Thus, malformed or otherwise low-reward responses remain trainable whenever their group contains a preference signal, whereas equal-reward groups and padding contribute neither tokens nor gradients. The estimator uses immediate action-level credit ($\gamma=0$) rather than propagating rewards across environment steps; any task-level lookahead is supplied by the search, constraint, or posterior oracle.}
{\rev Appendix~\ref{app:analysis} gives an idealized interpretation of verifier-grounded updates and their sensitivity to verifier disagreement.}

{\color{black}
\begin{table*}[t]
    \centering
    \newcommand{\std}[1]{{\scriptsize $\pm$ #1}}
    \caption{In-domain performance across the three training environments. Results are mean $\pm$ sample standard deviation over five {\rev evaluation} runs of 100 games. \textit{Oracle} (gray) directly executes the task oracle policy under the same evaluation protocol. Sokoban reports success rate (SR) only; Sudoku and Minesweeper additionally report completion rate (CR).}
    \label{tab:main_results}
    \setlength{\tabcolsep}{8pt}
    \begin{tabular}{c|c|cc|cc}
        \toprule
        \multirow{2}{*}{\textbf{Method}}
        & \textbf{Sokoban}
        & \multicolumn{2}{c|}{\textbf{Sudoku}}
        & \multicolumn{2}{c}{\textbf{Minesweeper}} \\
        & \scriptsize SR (\%)
        & \scriptsize SR (\%) & \scriptsize CR (\%)
        & \scriptsize SR (\%) & \scriptsize CR (\%) \\
        \midrule
        \rowcolor{gray!15}
        \textit{Oracle}
        & \textit{100.00 \std{0.00}}
        & \textit{100.00 \std{0.00}} & \textit{100.00 \std{0.00}}
        & \textit{78.60 \std{3.78}} & \textit{96.93 \std{0.77}} \\
        Base
        & 5.20 \std{3.35}
        & 0.00 \std{0.00} & 4.10 \std{0.39}
        & 0.20 \std{0.45} & 71.04 \std{2.00} \\
        GRPO
        & 12.20 \std{1.10}
        & 29.00 \std{3.46} & 39.03 \std{3.44}
        & 4.20 \std{1.30} & 73.49 \std{2.04} \\
        VinePPO
        & 6.80 \std{1.30}
        & 0.00 \std{0.00} & 2.90 \std{0.44}
        & 3.00 \std{2.24} & 72.75 \std{1.59} \\
        \rowcolor{blue!10}
        \textbf{VPR (Ours)}
        & \textbf{28.40 \std{2.79}}
        & \textbf{80.60 \std{4.16}} & \textbf{84.22 \std{3.15}}
        & \textbf{32.60 \std{5.03}} & \textbf{85.76 \std{0.98}} \\
        \bottomrule
    \end{tabular}
\end{table*}
}

\needspace{9\baselineskip}
\section{Experiments}
\label{sec:exp}
{\color{black}
\subsection{Experimental Setup}
\label{sec:exp:setup}

\textbf{Methods.}
We use Qwen3-4B~\citep{yang2025qwen3} as the base model and compare VPR with two reinforcement-learning baselines. GRPO optimizes outcome rewards through trajectory-level group normalization, while VinePPO~\citep{kazemnejad2025vineppo} uses Monte Carlo rollouts for step-level credit assignment. VPR combines the oracle-action-set verifier with state-group rollout and locally normalized optimization, as described in Sections~\ref{sec:method:preliminaries}--\ref{sec:method:optimization}.

\textbf{In-domain design.}
We first investigate whether repurposing an available task oracle as a source of action-level rewards improves learning relative to outcome-reward and rollout-based baselines. We train and evaluate each method separately on Sokoban, Sudoku, and Minesweeper, which correspond to the three oracle constructions introduced in Section~\ref{sec:method:vpr}. Each prompt contains only the current environment {\rev observation}, with $\texttt{history\_length}=0$. Sokoban uses $7\times7$ rooms with three boxes, Sudoku uses $9\times9$ puzzles with 40 blanks, and Minesweeper uses $5\times5$ boards with five mines.

For each update, VPR starts from 64 initial task instances and follows one committed trajectory from each instance. At every visited state, it samples $K=4$ candidate actions that form a state group. GRPO instead starts from 32 initial task instances and samples eight complete trajectories from each instance. The two methods therefore {\rev each generate 256 responses at the first decision of an update; total sequential rollout compute is not matched}. For VinePPO, we use three Monte Carlo continuations at each visited state. Appendix~\ref{app:impl} provides further training details.

We report success rate (SR), defined as the fraction of solved episodes. For Sudoku and Minesweeper, we additionally report completion rate (CR), which measures the fraction of correctly filled cells and revealed safe cells, respectively. We omit Sokoban CR because it is not comparable across the evaluated runs. All results are reported as the mean and sample standard deviation over five 100-episode {\rev evaluation} runs using the same five environment-seed sets across methods. {\rev These standard deviations describe evaluation variability for each checkpoint, not variability across independently trained models.}

\textbf{OOD transfer design.}
We next examine whether VPR supervision from {\rev multiple environments with task oracles improves performance on tasks beyond} the structured game environments. All models start from Qwen3-4B-Base. Base receives no reinforcement learning, and Math is trained only on DAPO-Math-17k~\citep{yu2025dapo}. Math+GRPO and Math+VPR augment the same math training with outcome-reward GRPO and VPR game interactions, respectively. Math isolates the effect of math reinforcement learning, while Math+GRPO {\rev provides a comparison with additional game experience but} without action-level process rewards. None of the downstream benchmarks is used for training.

All trained variants receive 100 updates. Each update uses 64 math prompts and samples eight responses per prompt. Math+VPR adds six Sokoban instances, eight Sudoku instances, and eighteen Minesweeper instances, sampling four candidates at every visited state. Math+GRPO instead uses three Sokoban instances, four Sudoku instances, and nine Minesweeper instances, sampling eight complete trajectories from each. These settings match {\rev the first-decision response counts} at 24 Sokoban responses, 32 Sudoku responses, and 72 Minesweeper responses per update{\rev{}, but not total sequential rollout compute or the visited-state distribution}.

\begin{table*}[t]
    \centering
    \scriptsize
    \newcommand{\std}[1]{{\fontsize{6pt}{7pt}\selectfont $\pm$ #1}}
    \newcommand{\gain}[1]{\,\textsuperscript{\textcolor{green!50!black}{\tiny +#1}}}
    \caption{Zero-shot general-reasoning performance from the Qwen3-4B-Base initialization. We report the mean $\pm$ sample standard deviation of pass@1 over $n$ independent evaluation runs, with $n$ shown under each benchmark; for $n=1$, only the score is reported. $\mathrm{Avg}$ gives equal weight to each benchmark, with green superscripts showing absolute gains over Base.}
    \label{tab:ood_math}
    \setlength{\tabcolsep}{5pt}
    \begin{tabular}{c|ccccccc|c}
        \toprule
        \textbf{Model}
        & \shortstack{\textbf{GSM8K}\\[-1pt]\tiny n=3}
        & \shortstack{\textbf{MATH-500}\\[-1pt]\tiny n=5}
        & \shortstack{\textbf{AIME24}\\[-1pt]\tiny n=32}
        & \shortstack{\textbf{AIME25}\\[-1pt]\tiny n=32}
        & \shortstack{\textbf{GPQA-D}\\[-1pt]\tiny n=5}
        & \shortstack{\textbf{BBH}\\[-1pt]\tiny n=1}
        & \shortstack{\textbf{MMLU-Pro}\\[-1pt]\tiny n=1}
        & \textbf{Avg} \\
        \midrule
        Base
        & 83.70 \std{0.95} & 67.20 \std{1.24} & 9.48 \std{4.24} & 7.19 \std{4.07}
        & 6.97 \std{1.73} & 14.71 & 17.01 & 29.47 \\
        Math
        & 91.38 \std{0.18} & 78.88 \std{0.23} & 19.06 \std{5.17} & 19.38 \std{3.92}
        & 35.45 \std{2.71} & 43.65 & 57.60 & 49.34\gain{19.87} \\
        Math+GRPO
        & 91.89 \std{0.45} & 81.20 \std{1.64} & 17.71 \std{4.19} & 17.19 \std{3.70}
        & 38.08 \std{1.50} & 54.23 & \textbf{58.91} & 51.32\gain{21.85} \\
        \rowcolor{blue!10}
        \textbf{Math+VPR}
        & \textbf{92.47 \std{0.46}} & \textbf{81.36 \std{1.40}} & \textbf{22.50 \std{3.59}} & \textbf{19.69 \std{3.91}}
        & \textbf{38.18 \std{2.54}} & \textbf{58.07} & 56.95 & \textbf{52.75}\gain{23.28} \\
        \bottomrule
    \end{tabular}
\end{table*}

\subsection{In-Domain Results}
\label{sec:exp:in_domain}

\textbf{Quantitative results.}
The Oracle row reports performance obtained by directly executing each environment's task oracle under the same evaluation protocol. Sokoban and Sudoku are solved deterministically, whereas the Minesweeper oracle must occasionally make minimum-risk guesses and therefore need not achieve perfect success. As shown in Table~\ref{tab:main_results}, VPR performs best among the learned policies on every reported metric. On Sokoban, it reaches $28.4\%$ SR, compared with $12.2\%$ for the strongest baseline. The difference is particularly pronounced on Sudoku, where VPR achieves $80.6\%$ SR compared with $29.0\%$ for GRPO. Minesweeper presents a complementary pattern: the base model attains substantial partial completion but almost never solves a board, whereas VPR reaches $32.6\%$ SR while all baselines remain below $5\%$. Together, these results demonstrate consistent in-domain improvements across the three forms of task structure considered here.

\begin{wrapfigure}{r}{0.50\textwidth}
\vspace{-11pt}
\centering
\includegraphics[width=\linewidth]{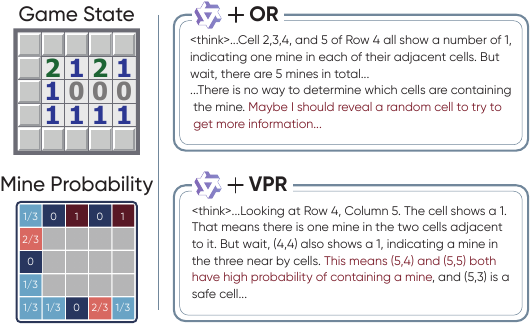}
\caption{{\rev Actual outcome-reward (OR) and VPR responses in an illustrative \textit{Minesweeper} case.}}
\label{fig:teaser}
\vspace{-8pt}
\end{wrapfigure}
\textbf{Pattern Analysis.}
Figure~\ref{fig:teaser} contrasts {\rev actual responses from an outcome-reward policy and a VPR policy to the same} illustrative \textit{Minesweeper} {\rev board}. {\rev The outcome-reward response proposes revealing a random cell, whereas the VPR response combines neighboring clues to identify $(5,3)$ as safe. The posterior verifier confirms that this cell has zero mine probability. This qualitative example illustrates how VPR-trained behavior can align with the local distinctions captured by the oracle, here recognizing a safe reveal rather than proposing an unnecessary guess.}

\WFclear
\subsection{OOD Generalization}
\label{sec:exp:ood}

Following the design above, we evaluate zero-shot general reasoning on seven benchmarks and agentic behavior on ALFWorld~\citep{shridhar2020alfworld} and WebShop~\citep{yao2022webshop}. We first report transfer to general reasoning and then to interactive agentic tasks.

\paragraph{General Reasoning.}
Table~\ref{tab:ood_math} shows that math reinforcement learning accounts for most of the improvement over the base model, while adding game interactions yields further average gains. Math+VPR achieves the highest macro average and the strongest observed mean on six of the seven benchmarks{\rev{} among the compared configurations}. {\rev Its average exceeds Math+GRPO by 1.43 percentage points, demonstrating an additional transfer benefit from the VPR training scheme.} {\rev Gains vary across tasks, and Math+GRPO remains stronger on MMLU-Pro.}

\begin{table}[t]
    \centering
    \small
    \newcommand{\std}[1]{{\scriptsize $\pm$ #1}}
    \newcommand{\gain}[1]{\,\textsuperscript{\textcolor{green!50!black}{\tiny +#1}}}
    \newcommand{\loss}[1]{\,\textsuperscript{\textcolor{red!70!black}{\tiny -#1}}}
    \caption{Zero-shot agentic OOD evaluation. ALFWorld reports success rate; WebShop reports task score and success rate. Superscripts show absolute changes from Base (green for gains, red for drops).}
    \label{tab:ood_agent}
    \setlength{\tabcolsep}{20pt}
    \begin{tabular}{c|c|cc}
        \toprule
        \multirow{2}{*}{\textbf{Model}}
        & \textbf{ALFWorld}
        & \multicolumn{2}{c}{\textbf{WebShop}} \\
        & \scriptsize SR (\%) & \scriptsize Score (\%) & \scriptsize SR (\%) \\
        \midrule
        Base & 1.34 \std{1.86} & 8.33 \std{6.23} & 0.93 \std{0.81} \\
        Math & 11.64 \std{2.40}\gain{10.30} & 21.79 \std{1.70}\gain{13.46} & 0.47 \std{0.12}\loss{0.46} \\
        Math+GRPO & 11.04 \std{1.70}\gain{9.70} & 32.76 \std{2.68}\gain{24.43} & 0.60 \std{0.20}\loss{0.33} \\
        \rowcolor{blue!10}
        \textbf{Math+VPR} & \textbf{16.12 \std{2.87}}\gain{14.78} & \textbf{42.01 \std{2.14}}\gain{33.68} & \textbf{1.33 \std{0.76}}\gain{0.40} \\
        \bottomrule
    \end{tabular}
\end{table}

\paragraph{Agentic Tasks.}
Table~\ref{tab:ood_agent} extends the observed transfer pattern to interactive benchmarks. Math+VPR achieves the strongest ALFWorld success rate and WebShop task score among the four training configurations. On WebShop, however, exact success remains low for every model despite substantial improvements in partial-credit score, indicating better progress toward task completion rather than consistently successful execution. Together with the general-reasoning results, these findings suggest that VPR supervision from multiple environments with task oracles can support transfer beyond the training interfaces.

\subsection{Ablation Studies}
\label{sec:exp:ablation}

\begin{wraptable}{r}{0.49\linewidth}
    \vspace{-10pt}
    \centering
    \small
    \newcommand{\std}[1]{{\scriptsize $\pm$ #1}}
    \caption{Sokoban ablations (SR, mean $\pm$ std over five 100-game {\rev evaluations}).}
    \label{tab:ablation_sokoban}
    \setlength{\tabcolsep}{18pt}
    \begin{tabular}{c|c}
        \toprule
        \textbf{Method} & \textbf{SR (\%)} \\
        \midrule
        Base & 5.20 \std{3.35} \\
        Vanilla VPR & 22.60 \std{4.77} \\
        \rowcolor{blue!10}
        \textbf{State-group VPR} & \textbf{28.40 \std{2.79}} \\
        20\% corruption & 14.60 \std{3.58} \\
        40\% corruption & 16.80 \std{3.42} \\
        \bottomrule
    \end{tabular}
    \vspace{-8pt}
\end{wraptable}

Table~\ref{tab:ablation_sokoban} first probes the combined benefit of oracle-guided state-group rollout and locally normalized optimization. State-group VPR outperforms vanilla VPR, consistent with the intended combination of same-state action comparison and guided exploration. {\rev This ablation evaluates their joint effect rather than isolating individual components.} For reward corruption rate $p$, each legal candidate independently has its oracle/non-oracle label flipped with probability $p$, swapping rewards $+1$ and $0$; invalid actions retain reward $-1$. Both corruption variants underperform clean state-group VPR, supporting the importance of verifier reliability{\rev{}, although performance does not decrease monotonically with the corruption rate}.

\subsection{Expert-as-Oracle VPR for General Tool Use}
\label{sec:exp:tau}

Finally, we study two questions beyond the structured game environments: whether VPR-style action-level supervision remains useful in more open-ended, general tool-use settings, and whether an expert reference policy can approximate a task oracle when no compact symbolic or algorithmic oracle is available. We train Qwen3-8B on the Airline and Retail domains of $\tau^2$-bench~\citep{barres2025tau2}. At each student-visited state, DeepSeek V4 Flash supplies reference actions used to score student tool calls by normalized exact matching and natural-language messages with a semantic matcher. The student trains with four-candidate state groups. Over three trials per task, $\mathrm{pass}^1$ reports mean single-trial success and $\mathrm{pass}^3$ reports the fraction of tasks solved in all three trials. Unlike pass@$k$, which counts success in at least one of $k$ attempts, $\mathrm{pass}^3$ measures repeatability. Appendix~\ref{app:tau} provides the full protocol and optimization details.

Unlike the {\rev symbolic or algorithmic} task oracles used in our game experiments, the expert reference policy provides an approximate and potentially incomplete action set. This setting therefore tests whether VPR can benefit from useful but imperfect action guidance in general tool use.

\begin{table*}[t]
    \centering
    \caption{Expert-as-oracle VPR results on $\tau^2$-bench. Each cell reports $\mathrm{pass}^1/\mathrm{pass}^3$ in percent: mean single-trial success and the fraction of tasks solved in all three trials, respectively.}
    \label{tab:tau2}
    \setlength{\tabcolsep}{8pt}
    \begin{tabular}{c|ccccc}
        \toprule
        \textbf{Method}
        & \shortstack{\textbf{Airline}\\\scriptsize train (27)}
        & \shortstack{\textbf{Airline}\\\scriptsize test (23)}
        & \shortstack{\textbf{Retail}\\\scriptsize train (61)}
        & \shortstack{\textbf{Retail}\\\scriptsize test (53)}
        & \shortstack{\textbf{Telecom}\\\scriptsize OOD (114)} \\
        \midrule
        Qwen3-8B (no RL)
        & 42.0 / 18.5 & 26.1 / 13.0 & 47.0 / 24.6 & 41.5 / 20.8 & 20.2 / \textbf{7.0} \\
        GRPO
        & 48.1 / 14.8 & 34.8 / 17.4 & 56.3 / 32.8 & 43.4 / 28.3 & 13.5 / 0.0 \\
        \rowcolor{blue!10}
        \textbf{VPR with expert}
        & \textbf{54.3 / 37.0} & \textbf{46.4 / 30.4} & \textbf{67.8 / 47.5} & \textbf{55.3 / 32.1} & \textbf{22.5} / 4.4 \\
        \bottomrule
    \end{tabular}
\end{table*}

Table~\ref{tab:tau2} shows that expert-guided VPR outperforms both the no-RL model and GRPO on $\mathrm{pass}^1$ and $\mathrm{pass}^3$ across the Airline and Retail training and held-out subsets. These results suggest that action-level VPR supervision remains useful in more open-ended tool-use interactions and that an expert can provide a practically useful approximate oracle within the trained domains. Transfer to the unseen Telecom domain is less consistent: VPR achieves the highest $\mathrm{pass}^1$, whereas the no-RL model retains the strongest $\mathrm{pass}^3$.

\subsection{Limitations}
\label{sec:exp:limitations}

The main limitation of VPR is its {\rev dependence on the availability and reliability of task oracles}. Such oracles are unavailable in many open-ended agentic tasks, and even when one exists, constructing or executing it may be costly. VPR also assumes parseable actions and sufficient task knowledge to identify oracle-preferred and legal actions. Moreover, state-group rollout shifts the distribution of visited states toward successors favored by the oracle{\rev{}, whose effect on learning may vary across tasks}. {\rev The expert-as-oracle extension depends on the quality and coverage of the expert's reference actions. Our comparisons address reward-based RL baselines; direct comparison with oracle imitation remains outside this study.} Extending VPR to less structured environments without a reliable source of action guidance remains an open challenge.
}

\section{Related Work}
\label{sec:related}
{\color{black}
\textbf{Reinforcement Learning from Verifiable Rewards.}
RLVR uses objective signals such as exact-answer checks, unit tests, symbolic solvers, and executable feedback in place of subjective preference judgments~\citep{ouyang2022training,uesato2022solving,le2022coderl,roziere2023code,pan2023logic,shao2024deepseekmath,guo2025deepseek}. Most existing approaches verify only a final answer or completed trajectory, leaving intermediate credit ambiguous in long-horizon interaction. VPR instead repurposes task-solving oracles based on search, constraints, or posterior inference as verifiers of parsed intermediate actions. When a reliable oracle {\rev can supply an action set at intermediate states}, this provides a dense alternative to outcome-reward training. VPR retains a PPO-style clipped objective while changing both the supervision source and rollout unit: candidate responses are compared locally at a shared state.

\textbf{Process Supervision and Credit Assignment.}
Process Reward Models assign feedback to intermediate steps using human or LLM annotations~\citep{lightman2024lets,wang-etal-2024-math,huang2024large,gou2024critic} or rollout-derived values~\citep{kazemnejad2025vineppo,yu-etal-2024-ovm}. Step-DPO and PRIME similarly obtain stepwise preferences or implicit rewards from error annotations and outcome-conditioned rollouts~\citep{lai2024stepdpo,cui2025prime}. VPR instead derives rewards by forming an oracle action set from task-solving guidance and scoring policy-generated actions against that set. VinePPO estimates an intermediate-state value using sampled continuations, whereas VPR compares actions sampled directly at the shared state. GiGPO~\citep{feng2025group} retrospectively forms micro-groups around repeated anchor states found in complete trajectories; VPR actively samples a state group at every visited state{\rev{} and scores its candidates with a task oracle}. VPR is also related to DAgger~\citep{ross2011dagger}, as both query oracle-derived information on {\rev states encountered during policy learning}, but VPR constructs local relative advantages rather than directly imitating one expert action. Its action-level rewards are not restricted to potential-based shaping~\citep{ng1999policy}; accordingly, we make no general claim of policy invariance.

\textbf{LLM Agents and Agentic Reinforcement Learning.}
LLM agents combine language-based reasoning with multi-turn environment interaction~\citep{xi2025rise,wang2024survey}. Inference-time methods such as ReAct, Reflexion, Tree of Thoughts, and LATS improve planning without updating model parameters~\citep{yao2023react,shinn2023reflexion,yao2024tree,zhou2024language}, whereas agentic RL learns from interactive experience~\citep{liu2024agentbench,chen2023fireact}. Recent work studies trajectory decomposition and turn-level credit in multi-turn MDPs~\citep{wang2025ragen,zeng2025turnlevel,luo2025agentlightning}. VPR focuses specifically on repurposing task-solving oracles as action-level process verifiers and coupling their rewards with {\rev active same-state sampling, local relative updates, and oracle-guided commits}. Our main evidence therefore comes from structured games with explicit task-solving oracles. On $\tau^2$-bench~\citep{barres2025tau2}, we further examine whether an expert reference policy can approximate an oracle for action-level supervision in general tool use.
}

\needspace{3\baselineskip}
\section{Conclusion}
\label{sec:conclusion}
{\color{black}
We presented Verifiable Process Rewards (VPR) for structured agentic reasoning problems that admit symbolic or algorithmic task-solving oracles. VPR repurposes each task oracle as an action-level process verifier by forming an oracle action set from the oracle-selected action and equivalent alternatives. {\rev The framework combines this reward interface with active same-state sampling, locally normalized updates, and oracle-guided commits, linking dense credit assignment with guided trajectory exploration.} Across Sokoban, Sudoku, and Minesweeper, VPR outperforms the base policy, outcome-reward GRPO, and VinePPO on all reported metrics.

Our mixed-training results further suggest that incorporating VPR supervision from multiple environments with task oracles can improve transfer beyond the training domains. With matched math exposure and {\rev first-decision game-response counts}, Math+VPR achieves the highest average score across the general-reasoning benchmarks and the strongest ALFWorld and WebShop results among the compared variants. On $\tau^2$-bench, {\rev expert-as-oracle VPR also improves performance on held-out Airline and Retail tasks, extending the approach to tool-use interactions with approximate oracle guidance}. Overall, these findings position VPR as a promising alternative to outcome-reward training when reliable task oracles are available. {\rev Extending these benefits to environments without reliable action guidance remains an open challenge.}
}

\bibliographystyle{plain}
\bibliography{reference}

\appendix
\section{\texorpdfstring{{\rev Analytical Interpretation}}{Analytical Interpretation}}
\label{app:analysis}

{\rev We use two standard policy-gradient observations to interpret verifier-grounded updates. Here, $\pi_\theta(a\mid s)$ denotes the action distribution induced by sampling a response from $\pi_\theta(\cdot\mid h(s))$ and applying the parser; conditioning this induced distribution on $s$ does not give the response policy access to hidden state information. Both observations concern a binary verifier and an unclipped objective under a fixed, $\theta$-independent state distribution. They do not analyze the combined effects of state-group collection, normalization, masking, PPO clipping, or changes in the visited-state distribution.}

\subsection{\texorpdfstring{{\rev Weighted-Imitation Interpretation}}{Weighted-Imitation Interpretation}}
\label{app:proof-vpr}

{\rev
{\textbf{Proposition 1 (First-order weighted-imitation interpretation).}
Consider a fixed state distribution $d(s)$ induced by the data-collection procedure and held independent of $\theta$. For the binary special case $\mathcal{V}(s,a)=\mathbb{I}(a\in\mathcal{A}_{\mathcal O}(s))$, the verifier objective $J_{\mathcal{V}}(\theta)=\mathbb{E}_{s\sim d,\,a\sim\pi_\theta}[\mathcal{V}(s,a)]$ has policy gradient}
\begin{equation}
    \nabla_\theta J_{\mathcal{V}}(\theta)
    =\mathbb{E}_{s\sim d,\,a\sim\pi_\theta}\!\left[\mathcal{V}(s,a)\nabla_\theta\log\pi_\theta(a\mid s)\right],
\end{equation}
{which is unchanged by subtracting an action-independent baseline. At $\theta=\theta_{\mathrm{old}}$, it also equals the gradient of a weighted log-likelihood objective over oracle-preferred sampled actions. This establishes a local, first-order connection to filtered imitation; it is not an equivalence between the complete VPR algorithm and behavioral cloning.}
}

\paragraph{\texorpdfstring{{\rev Proof.}}{Proof.}}

{\color{black}We prove the first-order policy-gradient interpretation in Proposition~1 under its stated idealized assumptions. The verifier objective is}
\begin{equation}
    J_{\mathcal{V}}(\theta)
    =
    \mathbb{E}_{s \sim d,\, a \sim \pi_\theta(\cdot \mid s)}
    \left[
    \mathcal{V}(s,a)
    \right],
\end{equation}
{\color{black}where $d(s)$ is a fixed state distribution and, for Proposition~1, $\mathcal{V}(s,a)=\mathbb{I}(a\in\mathcal A_{\mathcal O}(s))$ is the binary verifier induced by the oracle action set and is independent of $\theta$.} By the log-derivative identity,
\begin{equation}
\begin{aligned}
    \nabla_\theta J_{\mathcal{V}}(\theta)
    &=
    \nabla_\theta
    \sum_s d(s)
    \sum_a
    \pi_\theta(a \mid s)
    \mathcal{V}(s,a) \\
    &=
    \sum_s d(s)
    \sum_a
    \mathcal{V}(s,a)
    \nabla_\theta \pi_\theta(a \mid s) \\
    &=
    \sum_s d(s)
    \sum_a
    \pi_\theta(a \mid s)
    \mathcal{V}(s,a)
    \nabla_\theta \log \pi_\theta(a \mid s) \\
    &=
    \mathbb{E}_{s \sim d,\, a \sim \pi_\theta}
    \left[
    \mathcal{V}(s,a)
    \nabla_\theta \log \pi_\theta(a \mid s)
    \right].
\end{aligned}
\end{equation}
{\color{black}This establishes the policy-gradient identity.}

Next, for any action-independent baseline $b(s)$, we have
\begin{equation}
\begin{aligned}
    \mathbb{E}_{a \sim \pi_\theta(\cdot\mid s)}
    \left[
    b(s)\nabla_\theta \log \pi_\theta(a \mid s)
    \right]
    &=
    b(s)
    \sum_a
    \pi_\theta(a \mid s)
    \nabla_\theta \log \pi_\theta(a \mid s) \\
    &=
    b(s)
    \sum_a
    \nabla_\theta \pi_\theta(a \mid s) \\
    &=
    b(s)
    \nabla_\theta
    \sum_a
    \pi_\theta(a \mid s) \\
    &=
    b(s)
    \nabla_\theta 1
    =
    0.
\end{aligned}
\end{equation}
Therefore,
\begin{equation}
    \mathbb{E}_{a \sim \pi_\theta(\cdot\mid s)}
    \left[
    \left(\mathcal{V}(s,a)-b(s)\right)
    \nabla_\theta \log \pi_\theta(a \mid s)
    \right]
    =
    \mathbb{E}_{a \sim \pi_\theta(\cdot\mid s)}
    \left[
    \mathcal{V}(s,a)
    \nabla_\theta \log \pi_\theta(a \mid s)
    \right].
\end{equation}
{\color{black}Taking the expectation over $s\sim d$ yields the corresponding identity for the full gradient. Thus, subtracting an action-independent baseline can change variance but not the expected policy gradient.}

{\color{black}Finally, consider the following weighted log-likelihood objective:}
\begin{equation}
    L_{\mathrm{IL}}(\theta)
    =
    \mathbb{E}_{s \sim d,\, a \sim \pi_{\theta_{\mathrm{old}}}}
    \left[
    \mathcal{V}(s,a)
    \log \pi_\theta(a \mid s)
    \right].
\end{equation}
Its gradient is
\begin{equation}
    \nabla_\theta L_{\mathrm{IL}}(\theta)
    =
    \mathbb{E}_{s \sim d,\, a \sim \pi_{\theta_{\mathrm{old}}}}
    \left[
    \mathcal{V}(s,a)
    \nabla_\theta \log \pi_\theta(a \mid s)
    \right].
\end{equation}
Evaluating this gradient at $\theta=\theta_{\mathrm{old}}$ gives
\begin{equation}
    \left.
    \nabla_\theta L_{\mathrm{IL}}(\theta)
    \right|_{\theta=\theta_{\mathrm{old}}}
    =
    \mathbb{E}_{s \sim d,\, a \sim \pi_{\theta_{\mathrm{old}}}}
    \left[
    \mathcal{V}(s,a)
    \left.\nabla_\theta \log \pi_\theta(a \mid s)\right|_{\theta=\theta_{\mathrm{old}}}
    \right],
\end{equation}
which matches
\begin{equation}
    \left.
    \nabla_\theta J_{\mathcal{V}}(\theta)
    \right|_{\theta=\theta_{\mathrm{old}}}
    =
    \mathbb{E}_{s \sim d,\, a \sim \pi_{\theta_{\mathrm{old}}}}
    \left[
    \mathcal{V}(s,a)
    \left.\nabla_\theta \log \pi_\theta(a \mid s)\right|_{\theta=\theta_{\mathrm{old}}}
    \right],
\end{equation}
{\color{black}because $J_{\mathcal V}$ is evaluated on-policy at $\theta_{\mathrm{old}}$, where $\pi_\theta=\pi_{\theta_{\mathrm{old}}}$. Thus, at the behavior policy, the idealized verifier gradient agrees to first order with a weighted log-likelihood gradient over sampled actions in the oracle action set. This local identity does not imply that the full state-group VPR algorithm is equivalent to behavioral cloning.}

This completes the proof.

\subsection{\texorpdfstring{{\rev Sensitivity to Verifier Disagreement}}{Sensitivity to Verifier Disagreement}}
\label{app:proof-bias}

{\rev
{\textbf{Proposition 2 (Bias under binary verifier error).}
Consider the same idealized per-turn objective under a fixed, $\theta$-independent state distribution $d(s)$. Let the exact binary verifier be $\mathcal{V}^\star(s,a)=\mathbb{I}(a\in\mathcal{A}_{\mathcal O}(s))$. If a binary approximate verifier $\widehat{\mathcal{V}}$ disagrees with $\mathcal{V}^\star$ on a fraction $\bar\epsilon=\mathbb{E}_{s\sim d,a\sim\pi_\theta}[\mathbb{I}\{\widehat{\mathcal{V}}\ne\mathcal{V}^\star\}]$ of policy-sampled state--action pairs and $\|\nabla_\theta\log\pi_\theta(a\mid s)\|\le G$ almost surely, then the gradient bias satisfies}
\begin{equation}
    \big\|\widehat g(\theta)-g^\star(\theta)\big\|\le G\bar\epsilon.
\end{equation}
This is a linear upper bound on the idealized gradient bias in terms of verifier disagreement, not a claim that actual bias or final task performance varies linearly with the error rate.
}

\paragraph{\texorpdfstring{{\rev Proof.}}{Proof.}}

{\color{black}By the policy-gradient identity in Proposition~1, any binary verifier $\mathcal U\in\{0,1\}$ satisfies}
\begin{equation}
    \nabla_\theta J_{\mathcal U}(\theta)
    =
    \mathbb{E}_{s\sim d,\,a\sim\pi_\theta}
    \left[
    \mathcal U(s,a)\nabla_\theta\log\pi_\theta(a\mid s)
    \right].
\end{equation}
{\color{black}Subtracting the oracle-verifier gradient from the approximate-verifier gradient gives}
\begin{equation}
    \widehat g(\theta) - g^\star(\theta)
    =
    \mathbb{E}_{s\sim d,\,a\sim\pi_\theta}
    \left[
    \left(\widehat{\mathcal V}(s,a)-\mathcal V^\star(s,a)\right)
    \nabla_\theta\log\pi_\theta(a\mid s)
    \right].
\end{equation}
Since both $\widehat{\mathcal V}$ and $\mathcal V^\star$ are binary,
\begin{equation}
    \left|
    \widehat{\mathcal V}(s,a)-\mathcal V^\star(s,a)
    \right|
    =
    \mathbb{I}
    \left[
    \widehat{\mathcal V}(s,a)\neq\mathcal V^\star(s,a)
    \right].
\end{equation}
Using Jensen's inequality and the bounded-score assumption
$\|\nabla_\theta\log\pi_\theta(a\mid s)\|\le G$ almost surely, we obtain
\begin{equation}
\begin{aligned}
    \left\|
    \widehat g(\theta)-g^\star(\theta)
    \right\|
    &\le
    \mathbb{E}_{s\sim d,\,a\sim\pi_\theta}
    \left[
    \left|
    \widehat{\mathcal V}(s,a)-\mathcal V^\star(s,a)
    \right|
    \cdot
    \left\|
    \nabla_\theta\log\pi_\theta(a\mid s)
    \right\|
    \right] \\
    &\le
    G
    \mathbb{E}_{s\sim d,\,a\sim\pi_\theta}
    \left[
    \mathbb{I}
    \left[
    \widehat{\mathcal V}(s,a)\neq\mathcal V^\star(s,a)
    \right]
    \right] \\
    &=
    G\bar\epsilon.
\end{aligned}
\end{equation}
This completes the proof.

\paragraph{Remark.}
{\color{black}The bound is tight up to constants. If the approximate verifier disagrees with the oracle verifier on a measurable set of mass $\bar\epsilon$ and the score norm reaches $G$ in a coherent direction on that set, then the gradient difference can scale as $G\bar\epsilon$. The result applies to the idealized per-turn objective under a fixed state distribution. For an unnormalized sum over timesteps, the corresponding bound may additionally depend on the horizon.}

\section{Implementation Details}
\label{app:impl}

\paragraph{Framework and Software Stack.}
{\color{black}Our implementation is built on verl-agent~\citep{feng2025group}. Synchronous multi-turn generation uses vLLM~\citep{kwon2023efficient}, while policy and reference-model training use PyTorch Fully Sharded Data Parallel (FSDP). Each game environment exposes structured metadata to the rollout and trainer, including the parsed action, process reward, validity, terminal success, candidate rank, selected status, trajectory identifier, turn index, and \texttt{state\_group\_uid}. Sudoku generation uses GEM~\citep{liu2025gem}; Sokoban and Minesweeper use the corresponding deterministic local environment adapters and oracle implementations in our codebase.}

\paragraph{Training Settings.}
{\color{black}The standalone game experiments use Qwen3-4B. VPR starts 64 initial task instances per update and samples $K=4$ candidates at every visited state; its selected candidates form one committed trajectory per initial instance. The GRPO baseline starts 32 initial task instances and samples eight complete trajectories per instance. Both methods therefore generate 256 first-decision responses, although their later sequential rollout compute differs. Sokoban and Sudoku VPR train for 100 updates; Minesweeper VPR trains for 200. In all three games, an oracle action receives $+1$, another legal action receives $0$, and an illegal or unparsable action receives $-1$. The PPO mini-batch size is 32, the clipping radius is 0.2, and the actor learning rate is $10^{-6}$. The launch configurations enable an actor-side KL loss with coefficient $10^{-3}$ while disabling KL shaping of the environment reward. All games use immediate process credit ($\gamma=0$) without an additional outcome-reward bonus. The standalone estimator first centers rewards within each \texttt{state\_group\_uid} and then whitens all eligible centered advantages in the update batch. Equal-reward groups receive a zero loss mask. Rows copied only for distributed divisibility are marked by \texttt{is\_padding} and are excluded from statistics, evidence, metrics, advantages, and gradients.}

\paragraph{Evaluation Action Budgets.}
{\color{black}Evaluation allows at most 36 environment actions for Sokoban, 100 for Sudoku, and 25 for Minesweeper.}

{\color{black}The environment-level candidate count is independent of the inference engine's sequence count. All candidates are generated from a shared environment snapshot and may train the policy, but only a uniformly chosen maximum-reward candidate is executed. The selected successor state is then used to construct the next group.}

\paragraph{Generation Parameters.}
{\color{black}Game training uses thinking mode with temperature 1.0, top-$p=1.0$, and top-$k=-1$. Responses contain optional reasoning followed by exactly one parseable action in \texttt{<action>...\allowbreak</action>} tags. The Markovian game manager enforces \texttt{history\_length=0}: each turn exposes the current {\rev observation} but not the textual trajectory history.}

\paragraph{Hardware Configuration.}
{\color{black}The standalone game runs use one node with eight NVIDIA H100 80GB GPUs. vLLM uses tensor parallel size 2 and the trainer uses all eight GPUs.}

{\color{black}
\paragraph{Verifier Computation Cost.}
To distinguish oracle-query overhead from end-to-end training cost, we profile reference-action computation for each verifier using the same fixed CPU configuration. Table~\ref{tab:verifier_efficiency} reports the mean verifier time per environment step and its estimated wall-clock contribution to one RL update after accounting for the number of environment steps and parallel environment workers. These measurements isolate verifier execution and exclude LLM generation and policy optimization.

\begin{table*}[t]
    \color{black}
    \centering
    \caption{{\color{black}CPU profiling of verifier computation in the three training environments.}}
    \label{tab:verifier_efficiency}
    \setlength{\tabcolsep}{10pt}
    \begin{tabular}{lcc}
        \toprule
        \textbf{Environment / verifier}
        & \shortstack{\textbf{Mean verifier time}\\\textbf{per environment step}}
        & \shortstack{\textbf{Estimated verifier wall time}\\\textbf{per RL update}} \\
        \midrule
        Sokoban, BFS planner & 98.0 ms & 1.13 s \\
        Sudoku, constraint solver & 0.0650 ms & 2.41 ms \\
        Minesweeper, exact posterior & 0.458 ms & 2.05 ms \\
        \bottomrule
    \end{tabular}
\end{table*}
}

\section{Evaluation Details}
\label{app:eval}

{\rev The reported standard deviations describe repeated evaluations of each checkpoint, not independent training seeds.}

\paragraph{General Reasoning Benchmarks.}
{\color{black}We use EvalScope~\citep{evalscope_2024} v1.6.1 for GSM8K, MATH-500, AIME24/25, GPQA-Diamond, BBH, and MMLU-Pro. All evaluations are zero-shot, with thinking enabled, temperature 0.6, top-$p=0.95$, top-$k=20$, a 16,384-token output budget, and a 20,480-token model-length limit. For each benchmark, we report the mean and sample standard deviation of pass@1 over independent evaluation runs. GSM8K uses three runs; MATH-500 and GPQA-Diamond use five; AIME24 and AIME25 use 32; and BBH and MMLU-Pro use one. The average in Table~\ref{tab:ood_math} assigns equal weight to the seven benchmark means.}

\paragraph{Agentic Tasks.}
{\color{black}For interactive agentic tasks, we use verl-agent~\citep{feng2025group}.
\begin{itemize}
    \item \textbf{ALFWorld:} We report success rate over the 134-game OOD split using five sampling seeds.
    \item \textbf{WebShop:} We report task score and success rate on all 500 test instructions using three sampling seeds.
\end{itemize}
The same benchmark examples, prompts, parsers, environment seeds, and stopping rules are used for every checkpoint.}

{\color{black}
\section{Expert-as-Oracle \texorpdfstring{$\tau^2$}{tau2}-Bench Protocol}
\label{app:tau}

\paragraph{Task splits and qualification.}
Before RL, the expert runs four qualification trials on every candidate Airline and Retail training task; only tasks solved in all four trials enter training. This yields 27 Airline and 61 Retail training tasks. The testing subsets combine the official test tasks with training-split tasks rejected by the filter, giving 23 Airline and 53 Retail tasks. {\rev These held-out subsets are disjoint from the RL training pool but differ from the official test splits.} Telecom uses all 114 base-split tasks and is entirely OOD.

\paragraph{Expert-as-oracle process rewards.}
DeepSeek V4 Flash independently samples three reference actions at each student-visited state. Its input contains the domain policy and interaction history together with privileged hidden context: the task identifier, user scenario, evaluation criteria, and reference resolution actions. Student tool calls are compared with canonical expert tool actions by normalized exact matching; natural-language messages are evaluated by a batched LLM semantic matcher with individual retries on unresolved cases. A valid candidate matching at least one reference action receives $+1$, another valid candidate receives $0$, and an invalid or unparsable candidate receives $-1$. The expert is used only during VPR training and is removed for evaluation.

\paragraph{Optimization.}
VPR samples four student candidates per visited state, of which exactly one advances the environment. Unlike the standalone game estimator, the $\tau^2$ run uses DAPO with \texttt{state\_group\_uid} as the grouping key and group-standardized advantages; equal-reward groups are skipped. The outcome baseline uses vanilla rollout with trajectory-level outcome normalization and never invokes the expert or state-group path. Both methods run for 100 RL updates with eight initial tasks per update. VPR samples four candidates per visited state, whereas the outcome baseline samples four complete rollouts per initial task. Table~\ref{tab:tau2} reports the step-100 checkpoint for both GRPO and VPR.

\paragraph{Evaluation.}
Student sampling uses temperature 0.6, top-$p=0.95$, top-$k=20$, a 4,096-token response cap, and thinking mode. The user simulator is \texttt{openrouter/qwen/qwen3.6-27b} with temperature zero and reasoning disabled. Evaluation permits 30 student decisions and 10 invalid-action retries. Every task has three trials derived from seed base 300. We compute $\mathrm{pass}^1$ over all individual trials and $\mathrm{pass}^3$ as the fraction of tasks successful in all three trials; these consistency statistics differ from pass@$k$, which counts success in at least one of $k$ attempts. Scoring retains deterministic $\tau^2$ components but removes \texttt{NL\_ASSERTION}; no failed trial is excluded, and all reported domains have zero infrastructure failures. {\rev The task splits and scoring changes support comparisons among our evaluated models under this protocol, not direct comparison with scores from an unmodified official evaluation.}
}

\section{Game Observation and Prompt}

{\color{black}
\lstset{basicstyle=\ttfamily\scriptsize,aboveskip=2pt,belowskip=2pt,framesep=2pt}
\paragraph{\textbf{\textit{Sokoban}}}
Each Sokoban turn contains only the current board and the number of boxes. The action is one direction; prior reasoning and actions are not appended to the next prompt. Listing~\ref{lst:sokoban} shows the action-tag template with an example board.

\begin{lstlisting}[style=myprompt, basicstyle=\ttfamily\scriptsize, label={lst:sokoban}, caption={Markovian prompt for \textit{Sokoban}.}]
You are solving a Sokoban puzzle with 3 box(es).
Legend: # = wall, _ = floor, O = target,
X = box, P = player, * = box on target,
S = player on target

Current board:
#######
#_____#
#_P_XO#
#_____#
#_X_O_#
#_X_O_#
#######

Choose one move:
  up
  down
  left
  right

Format: <action>DIRECTION</action>
You may reason briefly in <think>...</think>
before your answer.
\end{lstlisting}
}

{\color{black}
\paragraph{\textbf{\textit{Sudoku}}}
The Sudoku observation contains the current grid and the coordinates of blank cells. Rows and columns are 1-indexed. The next prompt is rebuilt from the updated grid rather than from the dialogue history.

\begin{lstlisting}[style=myprompt, basicstyle=\ttfamily\scriptsize, label={lst:sudoku-initial}, caption={Markovian prompt for \textit{Sudoku}.}]
You are solving a Sudoku puzzle. Fill in the blank
cells (shown as .) using digits 1-9. Each row,
column, and 3x3 box must contain digits 1-9
exactly once.

Current grid:
4 . . | 9 5 . | 2 . 1
. . . | 3 6 . | . . .
. 6 . | . 8 4 | 9 5 3
...

Blank cells: (1,2), (1,3), (1,6), ...

Choose one blank cell and one digit to fill it.
Format: <action>ROW COL DIGIT</action>
(rows and columns are 1-indexed, e.g.
<action>3 5 7</action>)
You may reason briefly in <think>...</think>
before your answer.
\end{lstlisting}

\paragraph{\textbf{\textit{Minesweeper}}}
The Minesweeper observation contains only the current board, unrevealed cells, and flagged cells. Coordinates are 1-indexed and success means revealing every safe cell; flagging every mine is not required.

\begin{lstlisting}[style=myprompt, basicstyle=\ttfamily\scriptsize, label={lst:minesweeper-initial}, caption={Markovian prompt for \textit{Minesweeper}.}]
You are playing Minesweeper on a 5x5 board with
5 mines. Legend: . = hidden, F = flagged,
numbers = revealed (count of adjacent mines)

Current board:
. . . . .
. . 0 . .
. . 0 . .
. . . . .
. . . . .

Unrevealed cells: (1,1), (1,2), ...
Flagged cells: none

Choose one action:
  reveal ROW COL  - reveal a hidden cell
  flag ROW COL    - toggle a flag
Rows and columns are 1-indexed.

Format: <action>ACTION ROW COL</action>
You may reason briefly in <think>...</think>
before your answer.
\end{lstlisting}
}


\end{document}